\documentclass[runningheads]{llncs}

\usepackage[T1]{fontenc}

\usepackage{graphicx}
\usepackage{xcolor} 
\usepackage[most]{tcolorbox}
\usepackage{booktabs}
\usepackage{tikz}
\usepackage{tabularx}
\usepackage{float}
\usepackage{hyperref}

\tikzstyle{arrow} =[thick,->,>=stealth]

\begin{document}

\title{Exploring the Integration of Large Language Models into Automatic Speech Recognition Systems: An Empirical Study}

\author{Zeping Min\inst{1} \and
Jinbo Wang\inst{2}
}
\authorrunning{M. Author et al.}

\institute{Peking University \\ No.5 Yiheyuan Road, Haidian District, Beijing 100871, P.R.China \\
\email{zpm@pku.edu.cn}
\and
Peking University \\ No.5 Yiheyuan Road, Haidian District, Beijing 100871, P.R.China \\
\email{wangjinbo@stu.pku.edu.cn}\\
}
\maketitle             

\begin{abstract}
This paper explores the integration of Large Language Models (LLMs) into Automatic Speech Recognition (ASR) systems to improve transcription accuracy. The increasing sophistication of LLMs, with their in-context learning capabilities and instruction-following behavior, has drawn significant attention in the field of Natural Language Processing (NLP). Our primary focus is to investigate the potential of using an LLM's in-context learning capabilities to enhance the performance of ASR systems, which currently face challenges such as ambient noise, speaker accents, and complex linguistic contexts. We designed a study using the Aishell-1 and LibriSpeech datasets, with ChatGPT and GPT-4 serving as benchmarks for LLM capabilities. Unfortunately, our initial experiments did not yield promising results, indicating the complexity of leveraging LLM's in-context learning for ASR applications. Despite further exploration with varied settings and models, the corrected sentences from the LLMs frequently resulted in higher Word Error Rates (WER), demonstrating the limitations of LLMs in speech applications. This paper provides a detailed overview of these experiments, their results, and implications, establishing that using LLMs' in-context learning capabilities to correct potential errors in speech recognition transcriptions is still a challenging task at the current stage.

\end{abstract}

\section{Introduction}
\label{sec:introduction}

In today's era of cutting-edge technology, automatic speech recognition (ASR) systems have become an integral part. The advent of end-to-end ASR models, which are based on neural networks~\cite{dong2018speech,gulati2020conformer,chan2015listen,graves2006connectionist,chorowski2015attention,chan2016listen,graves2014towards,graves2013speech}, coupled with the rise of prominent toolkits such as ESPnet~\cite{watanabe2018espnet} and WeNet~\cite{yao2021wenet}, have spurred the progression of ASR technology. Nevertheless, ASR systems~\cite{tjandra2017listening,soltau2016neural,kim2017joint,sainath2019two,zhang2020unified,han2020contextnet,peng2022branchformer} occasionally yield inaccurate transcriptions, which can be attributed to ambient noise, speaker accents, and complex linguistic contexts, thus limiting their effectiveness.

Over the years, considerable emphasis has been placed on integrating a language model~\cite{kannan2018analysis,weiran22_interspeech} into the ASR decoding process. Language models have gradually evolved from statistical to neural. Recently, large language models (LLMs)~\cite{zhang2022opt,scao2022bloom,zeng2022glm,brown2020language,ouyang2022training,openai2023gpt4,touvron2023llama} have gained prominence due to their exceptional proficiency in a wide array of NLP tasks. Interestingly, when the parameter scale surpasses certain thresholds, these LLMs not only improve their performance but also exhibit unique features such as in-context learning and instruction following, thereby offering a novel interaction method.

Nevertheless, the attempts to leverage recent LLMs such as~\cite{ouyang2022training,openai2023gpt4,touvron2023llama} to boost ASR model performance are still in the nascent stages. This paper seeks to address this gap. Our primary focus is to explore the potential of employing an LLM's in-context learning capability to enhance ASR performance. Our methodology revolves around providing an appropriately designed instruction to the LLM, supplying it with the ASR transcriptions, and analyzing if it can rectify the mistakes.

We employed the Aishell-1 and LibriSpeech datasets for our experiments and selected well-known LLM benchmarks, such as ChatGPT and GPT-4, which are generally considered superior to other LLMs for their comprehensive capabilities. We concentrated on the potential of GPT-3.5 and GPT-4 to correct possible errors in speech recognition transcriptions.\footnote{Although we conducted preliminary trials with models like llama, opt, bloom, etc., these models often produced puzzling outputs and rarely yielded anticipated transcription corrections.} Our initial experiments with the GPT-3.5-16k (GPT-3.5-turbo-16k-0613) model, in a one-shot learning scenario, did not yield lower WER.

Consequently, we undertook further investigation using diverse settings, including variations in the LLM model (GPT-3.5-turbo-4k-0301, GPT-3.5-turbo-4k-0613, GPT-4-0613), modification of instructions, increasing the number of attempts (1, 3, and 5), and varying the number of examples supplied to the model (1, 3, and 5-shot settings). This paper presents a thorough discussion of these experiments, their outcomes, and our insights. Regrettably, the findings of these experiments suggest that, at the present stage, directly employing the in-context learning ability of LLMs to correct potential errors in speech recognition transcriptions is extremely challenging and often leads to a \emph{higher WER}. This may be due to the lack of ability of LLMs in speech transcription.

\label{sec:contributions}
This study contributes to the field in three ways:

\begin{enumerate}
\item \textbf{Exploration of LLMs for ASR Improvement:} We explore the potential of large language models (LLMs), particularly focusing on GPT-3.5 and GPT-4, to improve automatic speech recognition (ASR) performance by their in-context learning ability. This is an emerging area of research, and our work contributes to its early development.

\item \textbf{Comprehensive Experiments Across Various Settings:} We conduct comprehensive experiments using the Aishell-1 and LibriSpeech datasets and analyze the effect of multiple variables, including different LLM models, alterations in instructions, varying numbers of attempts, and the number of examples provided to the model. Our work contributes valuable insights into the capabilities and limitations of LLMs in the context of ASR.

\item \textbf{Evaluation of the Performance:} Regrettably, our findings indicate that leveraging the in-context learning ability of LLMs to correct potential errors in speech recognition transcriptions often leads to a higher word error rate (WER). This critical evaluation underscores the current limitations of directly applying LLMs in the field of ASR, thereby identifying an important area for future research and improvement.
\end{enumerate}

\section{Related Work}
\label{sec:related_work}

The use of large language models (LLMs) to enhance the performance of automatic speech recognition (ASR) models has been the subject of numerous past studies \cite{udagawa2022effect,shin2019effective,chiu2021innovative,xu2022rescorebert,futami2020distilling,kubo2022knowledge}. These works have explored various strategies, including distillation methods \cite{futami2020distilling,kubo2022knowledge} and rescoring methods \cite{udagawa2022effect,shin2019effective,chiu2021innovative,xu2022rescorebert}.

In the distillation approach, for instance, \cite{futami2020distilling} employed BERT in the distillation approach to produce soft labels for training ASR models. \cite{kubo2022knowledge} strived to convey the semantic knowledge that resides within the embedding vectors.

For rescoring methods, \cite{shin2019effective} adapted BERT to the task of n-best list rescoring. \cite{chiu2021innovative} redefined N-best hypothesis reranking as a prediction problem. \cite{xu2022rescorebert} attempted to train a BERT-based rescoring model with MWER loss. \cite{udagawa2022effect} amalgamated LLM rescoring with the Conformer-Transducer model.

However, the majority of these studies have employed earlier LLMs, such as BERT \cite{devlin2018bert}. Given the recent explosive progress in the LLM field, leading to models with significantly more potent NLP abilities, such as ChatGPT, it becomes crucial to investigate their potential to boost ASR performance. Although these newer LLMs have considerably more model parameters, which can pose challenges to traditional distillation and rescoring methods, they also possess a crucial capability, in-context learning, which opens up new avenues for their application. 

\section{Methodology}
\label{sec:methodology}

\begin{figure}
    \centering
    \includegraphics[scale=0.5]{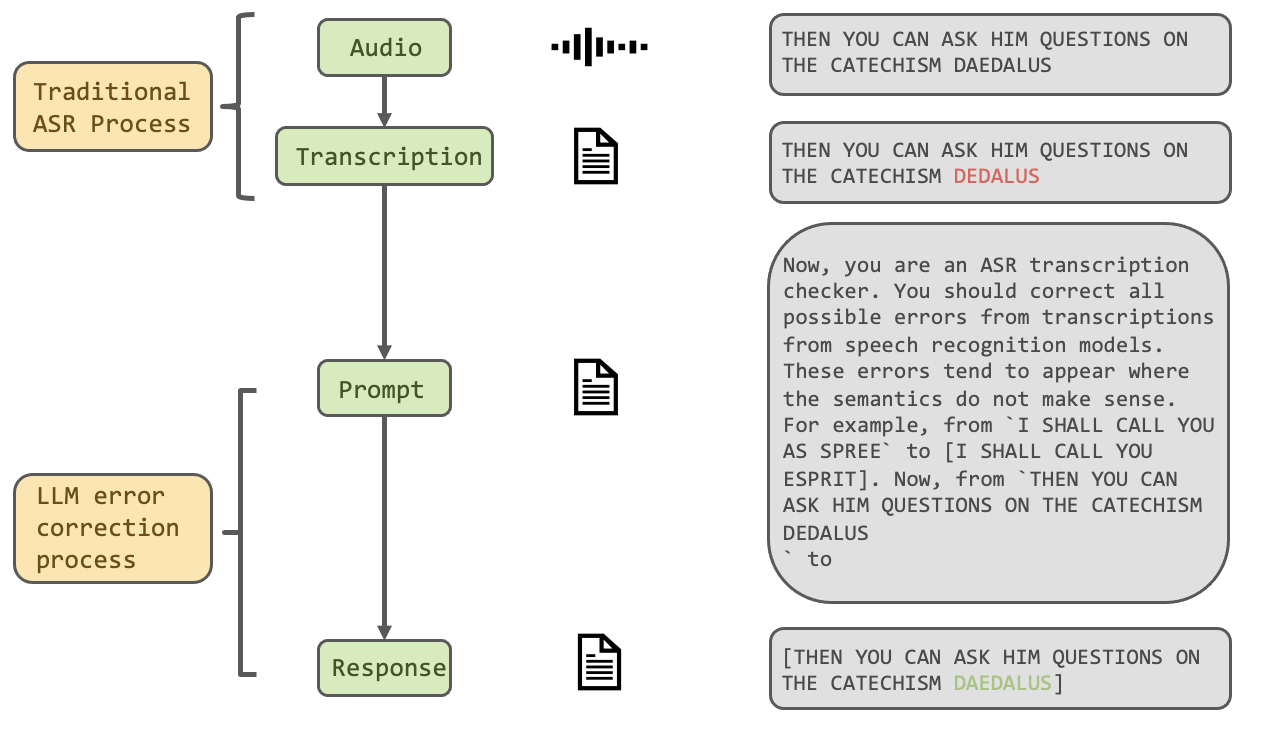}
    \caption{Overview of the methodology leveraging the in-context learning capability of large language models (LLMs) for potential correction of errors in automatic speech recognition (ASR) transcriptions.}
    \label{fig:enter-label}
\end{figure}

Our approach leverages the in-context learning abilities of LLMs. We supply the LLMs with the ASR transcription results and a suitable instruction to potentially correct errors. The process can be formalized as:
$$y=LLM(I,(x_1,y_1),(x_2,y_2),...,(x_k,y_k),x)$$
where $x$ represents the ASR transcription result, and $y$ is the correct transcription. The pairs $(x_i,y_i)_{i=1}^k$ are the $k$ examples given to the LLM, and $I$ is the instruction provided to the LLM. The prompt is represented by $(I,(x_1,y_1),(x_2,y_2)\\,...,(x_k,y_k),x)$. The entire process is visually illustrated in Figure \ref{fig:enter-label}.

We conducted thorough experimentation, varying GPT versions, the design of the instruction, and the number of examples $k$ provided to GPT, in order to assess the potential of using Large Language Models (LLMs) to improve Automatic Speech Recognition (ASR) performance. We tested three versions of GPT-3.5, as well as the high-performing GPT-4. We used four carefully crafted instructions and varied the number of examples, where $k=1,2,3$, supplied to the LLM.

Unfortunately, we found that directly applying the in-context learning capabilities of the LLM models for improving ASR transcriptions presents a significant challenge, and often leads to a higher Word Error Rate (WER). We further experimented with multiple attempts at sentence-level corrections. That is, for each transcription sentence $x$, the LLM generates multiple corrected outputs, and the final corrected result of the transcription sentence $x$ is chosen as the output with the least WER.\footnote{Selecting the output with the lowest WER is not practical in real-world scenarios, as we cannot know the actual transcription $y$. Nonetheless, this technique aids in comprehending the limitations of using LLM's in-context learning capabilities for enhancing ASR transcriptions.} Regrettably, even with multiple attempts, the corrected output from the LLM still results in a higher WER, further substantiating the challenges associated with directly leveraging the LLM's in-context learning capabilities for enhancing ASR transcriptions.

\section{Experiments}
\label{sec:experiments}
\setlength{\tabcolsep}{13pt}
\subsection{Setup}
\subsubsection{Dataset}
For our investigation, we selected two distinct datasets to evaluate the efficacy of utilizing advanced LLMs to improve ASR performance: the Aishell-1 dataset for the Chinese language and the LibriSpeech dataset for the English language. These datasets are greatly appreciated in the ASR research field, serving as standard benchmarks for numerous studies and methodologies.

The Aishell-1 \cite{bu2017aishell} dataset has a total duration of 178 hours, with the precision of manual transcriptions exceeding 95\%. The dataset is meticulously organized into training, development, and testing subsets.

In contrast, the LibriSpeech \cite{panayotov2015librispeech} dataset comprises approximately 1000 hours of English speech sampled at 16kHz. The content is extracted from audiobooks as a part of the LibriVox project. Similar to Aishell-1, LibriSpeech is also partitioned into subsets for training, development, and testing. Furthermore, each subset is classified into two groups based on data quality: clean and other.

\subsubsection{ASR Model}

To ensure the applicability of our experimental results, we utilized a state-of-the-art hybrid CTC/attention architecture, highly regarded in the field of speech recognition. We employed pretrained weights provided by the Wenet \cite{yao2021wenet} speech community.

The ASR model, trained on the Aishell-1 dataset, includes an encoder set up with a swish activation function, four attention heads, and 2048 linear units. The model employs an 8-kernel CNN module with layer normalization, and normalizes the input layer before activation. The encoder consists of 12 blocks, has an output size of 256, and uses gradient clipping (value=5) to prevent gradient explosions. The model leverages the Adam optimizer with a learning rate of 0.001 and a warm-up learning rate scheduler that escalates the learning rate for the initial 25,000 steps.

The ASR model, trained on the Librispeech dataset, implements a bitransformer decoder and a conformer encoder. The encoder follows the same configuration as that of the Aishell-1 model. The decoder incorporates four attention heads, with a dropout rate of 0.1. The model adheres to the same optimization and learning rate strategies as the Aishell-1 model.

\setlength{\tabcolsep}{10pt}
\begin{table}[ht]
\centering
\caption{Instructions for ASR transcription correction.}
\label{tab:instructions}
\begin{tabular}{cp{9cm}}
\toprule
\textbf{Instruction ID} & \textbf{Description} \\
\midrule
Instruction 1 & Correct the following transcription from speech recognition. \\
\midrule
Instruction 2 & Now, you are an ASR transcription checker. You should correct all possible errors from transcriptions from speech recognition models. These errors tend to appear where the semantics do not make sense. \\
\midrule
Instruction 3 & I have recently started using a speech recognition model to recognize some speeches. Of course, these recognition results may contain some errors. Now, you are an ASR transcription checker, and I need your help to correct these potential mistakes. You should correct all possible errors from transcriptions from speech recognition models. These errors often occur where the semantics do not make sense and can be categorized into three types: substitution, insertion, and deletion. \\
\midrule
Instruction 4 & I have recently been using a speech recognition model to recognize some speeches. Naturally, these recognition results may contain errors. You are now an ASR transcription checker, and I require your assistance to correct these potential mistakes. Correct all possible errors from transcriptions provided by the speech recognition models. These errors typically appear where the semantics don't make sense and can be divided into three types: substitution, insertion, and deletion. Please use '[]' to enclose your final corrected sentences. \\
\bottomrule
\end{tabular}
\end{table}

\subsubsection{LLM}
For the LLM models, considering that ChatGPT and GPT-4 are recognized benchmarks, we inspected three versions from ChatGPT (GPT-3.5-turbo-4k-0301, GPT-3.5-turbo-4k-0613, GPT-3.5-turbo-16k-0613) and GPT-4 (GPT-4-0613). While other LLMs such as Llama, Opt, and Bloom claim to equal or outperform ChatGPT in certain aspects, they generally fall behind ChatGPT, and even more so GPT-4, in terms of overall competency for generic tasks. For the instruction $I$, we tested four variations, as detailed in Table \ref{tab:instructions}.

Concerning the examples input to the LLM, we assessed 1-shot, 2-shot, and 3-shot scenarios. For the number of attempts, we explored situations with 1-attempt, 3-attempts, and 5-attempts.

Since the LLM output may contain some irrelevant content with ASR transcription, for testing convenience, we devised a method suite to extract transcriptions from LLM output. Specifically, from the prompt perspective, we tell the model to enclosed the corrected transcription in '[ ]', either by presenting the example to the model or directing the model in the instruction. After the LLM generates the text, we initially extract the text within '[ ]' from the LLM output text. In the following step, we eliminate all the punctuation within it. This is because the ground truth transcription provided by the Aishell-1 dataset and Librispeech dataset does not include punctuation.

\subsection{Results}
In our preliminary experiments, we established a baseline using the GPT-3.5 (GPT-3.5-turbo-16k-0613 version) model. We employed the Instruction 1: \emph{Correct the following transcription from speech recognition.} We employed a single attempt with one-shot learning. The outcomes of these initial tests, as shown in Table~\ref{tab:baseline}, were unsatisfactory.

\begin{table}[H]
\centering
\caption{WER (\%) results using GPT-3.5-16k-0613 for ASR transcription correction with one-shot learning.}
\label{tab:baseline}
\begin{tabular}{cccc}
\toprule
 & Aishell-1 & \multicolumn{2}{c}{LibriSpeech} \\
 \cmidrule(lr){3-4}
 & & Clean & Other \\
\midrule
with LLM & 12.36& 47.93& 51.25\\
without LLM & 4.73 & 3.35 & 8.77 \\
\bottomrule
\end{tabular}
\end{table}

\begin{figure}[H]
    \centering
    \includegraphics[scale=0.6]{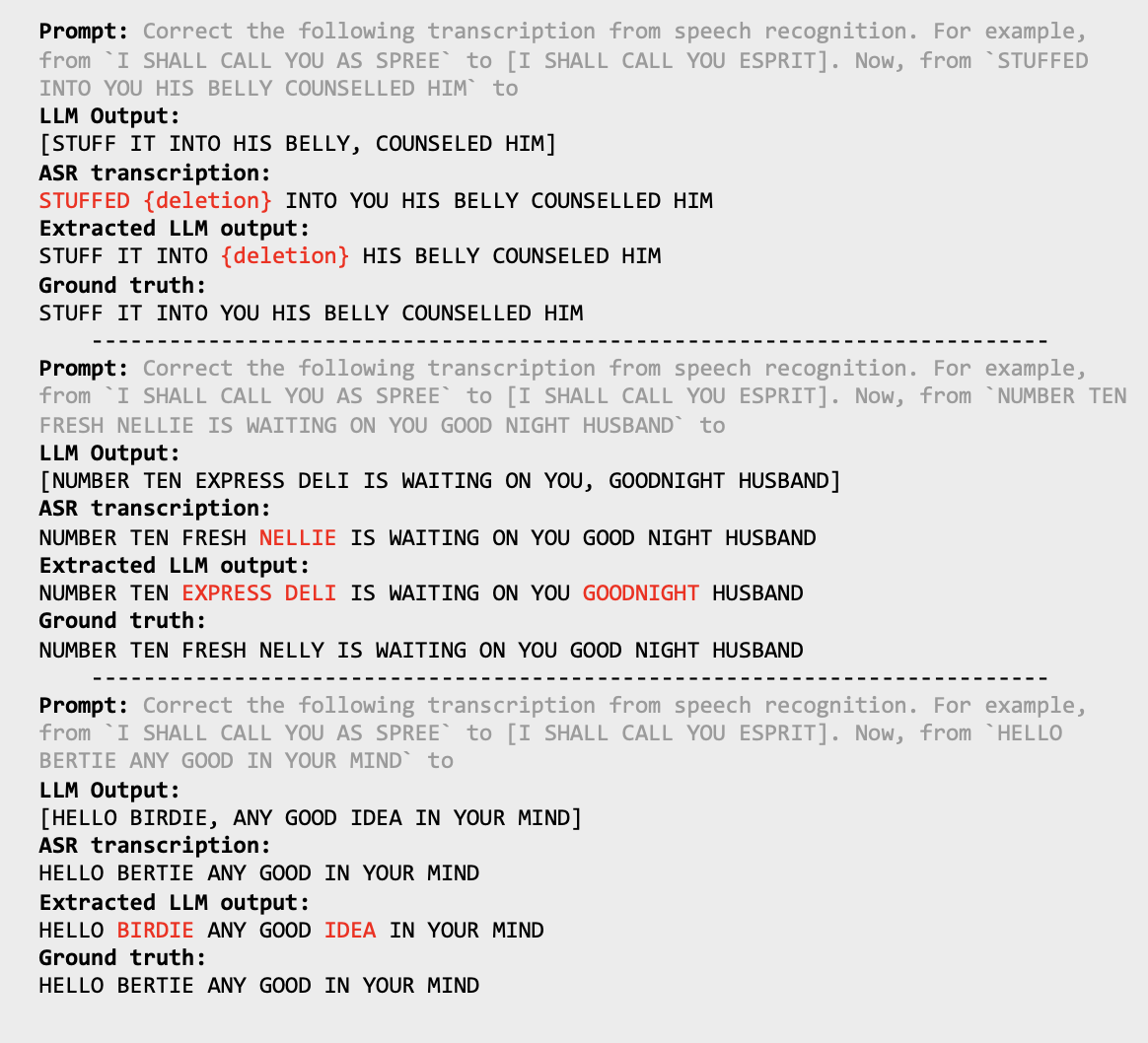}
    \caption{Illustrative examples of the LLM's challenges in interpreting and correcting ASR transcriptions from LibriSpeech Clean dataset.}
    \label{fig:exam-label}
\end{figure}

Furthermore, we provide several examples from LibriSpeech Clean in Figure \ref{fig:exam-label}. These instances underscore the challenges the LLM encounters when interpreting and correcting ASR transcriptions, which result in unsatisfactory performance for our task. In the first example, the original transcription read, "STUFFED INTO YOU HIS BELLY COUNSELLED HIM". Yet, the LLM amended "YOU" to "HIS" in the corrected transcription, deviating from the ground truth. In the second instance, the original transcript stated, "NUMBER TEN FRESH NELLIE IS WAITING ON YOU GOOD NIGHT HUSBAND". However, the LLM altered "FRESH NELLIE" to "EXPRESS DELI", thereby significantly modifying the intended meaning. In the third case, the initial transcription was "HELLO BERTIE ANY GOOD IN YOUR MIND". However, the LLM misinterpreted "BERTIE" as "BIRDIE" and superfluously appended "IDEA" to the corrected transcription.

\subsubsection{Results with Different LLM Models}
\label{subsec:expLLM}

Initially, we varied the LLM models utilized in our experiments, considering three different versions: GPT-3.5-turbo-4k-0301, GPT-3.5-turbo-4k-0613, and GPT-3.5-turbo-16k-0613 models. The results are consolidated in Table~\ref{tab:llm}. 

The observations from Table~\ref{tab:llm} demonstrate that all LLM models have a higher Word Error Rate (WER) than the scenario without the utilization of an LLM. This finding suggests that while LLM models exhibit potential for a broad range of NLP applications, their application for error correction in ASR transcriptions still requires refinement. Notably, the WER for all LLM models in the Aishell-1 dataset is significantly higher than the WER in the scenario without an LLM. A similar pattern is evident in the LibriSpeech dataset, for both clean and other data. Furthermore, the performance of GPT-3.5-turbo-4k-0613 and GPT-3.5-turbo-16k-0613 models is markedly better than that of the GPT-3.5-turbo-4k-0301 model. This disparity could be due to the enhancements in the GPT-3.5 model.

\begin{table}[ht]
\centering
\caption{WER (\%) performance comparison of different LLM models on ASR transcription error correction.}
\label{tab:llm}
\begin{tabular}{cccc}
\toprule
 & Aishell-1 & \multicolumn{2}{c}{LibriSpeech} \\
 \cmidrule(lr){3-4}
 & & Clean & Other \\
\midrule
GPT-3.5-turbo-4k-0301 & 16.05& 57.83& 51.20\\
GPT-3.5-turbo-4k-0613 & 12.32& 47.57& 51.10\\
GPT-3.5-turbo-16k-0613 & 12.36& 47.93& 51.25\\
without LLM & 4.73 & 3.35 & 8.77 \\
\bottomrule
\end{tabular}
\end{table}

\subsubsection{Results with Varying Instructions}
\label{subsec:expPrompts}

Next, we carried out a series of experiments using a variety of instructions. We precisely constructed four different types of instructions, which are displayed in Table~\ref{tab:instructions}. These instructions gradually provided more specific guidance for the task. We utilized the GPT-3.5-turbo-16k-0613 model for this purpose. The outcomes for the different instructions are tabulated in Table~\ref{tab:prompts}.

Furthermore, we tested varying instructions with two different models, specifically GPT-3.5-turbo-4k-0301 and GPT-3.5-turbo-4k-0613, the results of which are included in Appendix~\ref{app_prompt}. Our findings suggested that supplying detailed instructions to the Language Model (LLM) improves its performance. However, even with extremely detailed instructions, the LLM model does not demonstrate adequate performance in the task of rectifying errors in speech recognition transcriptions. That is to say, the Word Error Rate (WER) escalates after correction.\footnote{One might think that more detailed instructions could lead to better performance. This is indeed possible. In fact, we have exhaustively tried a lot of other instructions, but we have not observed a lower WER after corrections made by the LLM.}

\begin{table}[ht]
\centering
\caption{WER (\%) comparison for varying instructions with the GPT-3.5-turbo-16k-0613 model.}
\label{tab:prompts}
\begin{tabular}{cccc}
\toprule
 & Aishell-1 & \multicolumn{2}{c}{LibriSpeech} \\
 \cmidrule(lr){3-4}
 & & Clean & Other \\
\midrule
Instruction 1 & 12.36& 47.93& 51.25\\
Instruction 2 & 34.08& 48.58& 64.60\\
Instruction 3 & 22.32& 37.21& 48.14\\
Instruction 4 & 12.22& 23.93& 17.17\\
without LLM & 4.73 & 3.35 & 8.77 \\
\bottomrule
\end{tabular}
\end{table}

\subsubsection{Results with Varying Shots}
\label{subsec:expShots}

Subsequently, we explored the impact of varying the number of examples given to the model, using 1-shot, 2-shot, and 3-shot configurations. We utilized the GPT-3.5-turbo-16k-0613 model. For instructions, we employed the most detailed Instruction 4, which was proven to yield superior results in Subsection \ref{subsec:expPrompts}. The results are encapsulated in Table~\ref{tab:shots}. Moreover, we also tested Instructions 1, 2, and 3, with the outcomes detailed in Appendix \ref{app_shot}. Our observations revealed that providing the model with more examples led to enhanced performance, aligning with findings observed in many NLP tasks involving LLM. However, for the 1-shot, 2-shot, and 3-shot scenarios we experimented with, none of them resulted in a satisfactory WER, indicating an increase in errors post-correction. This is consistent with our previous observation that more progress is needed to harness LLMs effectively for ASR transcription error correction.

\begin{table}[ht]
\centering
\caption{WER (\%) comparison for varying shots with Instruction 4 and the GPT-3.5-turbo-16k-0613 model.}
\label{tab:shots}
\begin{tabular}{cccc}
\toprule
 & Aishell-1 & \multicolumn{2}{c}{LibriSpeech} \\
 \cmidrule(lr){3-4}
 & & Clean & Other \\
\midrule
1-shot & 12.22& 23.93& 17.17\\
2-shot & 14.19& 23.38& 17.68\\
3-shot & 12.71& 22.68& 17.43\\
without LLM & 4.73 & 3.35 & 8.77 \\
\bottomrule
\end{tabular}
\end{table}

\subsubsection{Results with Varying Attempts}
\label{subsec:expAttempts}

In the previous subsections, we established that the performance of Language Model Large (LLMs) in correcting errors in Automatic Speech Recognition (ASR) transcriptions is currently unsatisfactory, as corrections generally increase the number of errors. To deepen our understanding of the LLM's limitations in error correction for ASR transcriptions, we conducted further tests allowing the model multiple attempts. Specifically, for each transcription sentence $x$, the LLM generates multiple corrected outputs, and the final corrected result of the transcription sentence $x$ is chosen as the output with the least Word Error Rate (WER). In practical applications, choosing the output with the lowest WER is not feasible, as the correct transcription $y$ is unknown. Nevertheless, this approach aids in elucidating the constraints of leveraging LLM's in-context learning capabilities for ASR transcription enhancement. We present the results for 1, 3, and 5 attempts in Table~\ref{tab:attempts}. We utilized the GPT-3.5-turbo-16k-0613 model. For instructions, we employed the most effective Instruction 4 from Subsection \ref{subsec:expPrompts}. Additionally, in the prompt, we provided the model with three examples, that is, the 3-shot setup. Refer to Table~\ref{tab:attempts} for the experimental results. We discovered that even with up to five trials allowed, and the optimal result taken on a per-sentence basis, the outputs of the LLM still introduce more errors.

\begin{table}[ht]
\centering
\caption{WER (\%) comparison for varying attempts with Instruction 4.}
\label{tab:attempts}
\begin{tabular}{cccc}
\toprule
 & Aishell-1 & \multicolumn{2}{c}{LibriSpeech} \\
 \cmidrule(lr){3-4}
 & & Clean & Other \\
\midrule
1 Attempt & 12.71& 22.68& 17.43\\
3 Attempts & 6.81& 17.50& 12.54\\
5 Attempts & 5.77& 15.90& 11.49\\
without LLM & 4.73 & 3.35 & 8.77 \\
\bottomrule
\end{tabular}
\end{table}

\subsubsection{GPT4 Experimentations}
\setlength{\tabcolsep}{8pt}

We further extended our study to include the latest GPT4 model, currently deemed the most advanced. Due to the high computational demand and RPM restrictions of GPT4, we limited our testing to the LibriSpeech clean test set. We conducted tests using a one-shot setting for the four detailed instructions provided in Table \ref{tab:instructions}. The outcomes are encapsulated in Table \ref{tab:gpt4}. Our findings indicated that, despite employing the state-of-the-art GPT4 model, the ASR transcriptions corrected with LLM still yielded a higher number of errors.

\begin{table}[ht]
\centering
\caption{WER (\%) results with the GPT4 model for the LibriSpeech clean test set.}
\label{tab:gpt4}
\begin{tabular}{ccccc}
\toprule
Instruction 1 & Instruction 2 & Instruction 3 & Instruction 4 & Without LLM \\
\midrule
28.97& 23.91& 16.76& 14.90 & 3.35\\
\bottomrule
\end{tabular}
\end{table}

\section{Conclusion}
\label{sec:conclusion}

This paper has provided an exploratory study on the potential of Large Language Models (LLMs) to rectify errors in Automatic Speech Recognition (ASR) transcriptions. Our research focused on employing renowned LLM benchmarks such as GPT-3.5 and GPT-4, which are known for their extensive capabilities. Our experimental studies included a diverse range of settings, variations in the LLM models, changes in instructions, and a varied number of attempts and examples provided to the model.

Despite these extensive explorations, the results were less than satisfactory. In many cases, sentences corrected by LLMs resulted in higher Word Error Rates (WERs), thus revealing the limitations of LLMs in speech applications. This outcome points to the significant challenges in directly leveraging the in-context learning abilities of LLMs to improve ASR transcriptions.

These findings do not imply that the application of LLMs in ASR technology should be dismissed. On the contrary, they suggest that further research and development are required to optimize the use of LLMs in this area. As LLMs continue to evolve, their capabilities might be harnessed more effectively in the future to overcome the challenges identified in this study.

In conclusion, while the use of LLMs for enhancing ASR performance is in its early stages, the potential for improvement exists. This study hopes to inspire further research in this field, with the aim of refining and improving the application of LLMs in ASR technology.

\bibliographystyle{splncs04}
\bibliography{mybibliography}

\section*{Appendix}
\subsection*{Results with Varying Instructions}
\label{app_prompt}

We conducted experiments with various instructions. Four distinct types of instructions were meticulously designed, as depicted in Table \ref{tab:instructions}. These instructions progressively provided more detailed task directives. The experimental results for the GPT-3.5-turbo-4k-0301 and GPT-3.5-turbo-4k-0613 models, under the conditions of these four instructions, are presented in Table \ref{tab:app_prompts_1} and Table \ref{tab:app_prompts_2}, respectively. Our findings suggest that supplying the LLM model with detailed instructions aids in achieving enhanced performance. However, even with highly detailed instructions, the LLM model's performance in the task of correcting speech recognition transcription errors is not satisfactory. That is to say, the Word Error Rate (WER) increases post-correction.

\begin{table}[H]
\centering
\caption{WER comparison for varying instructions with the GPT-3.5-turbo-4k-0301 model.}
\label{tab:app_prompts_1}
\begin{tabular}{cccc}
\toprule
 & Aishell-1 & \multicolumn{2}{c}{LibriSpeech} \\
 \cmidrule(lr){3-4}
 & & Clean & Other \\
\midrule
Instruction 1 & 16.05& 57.83& 51.20\\
Instruction 2 & 16.81& 30.85& 36.99\\
Instruction 3 & 14.12& 24.42& 26.19\\
Instruction 4 & 14.16& 25.56& 20.26\\
without LLM & 4.73 & 3.35 & 8.77 \\
\bottomrule
\end{tabular}
\end{table}

\begin{table}[H]
\centering
\caption{WER comparison for varying instructions with the GPT-3.5-turbo-4k-0613 model.}
\label{tab:app_prompts_2}
\begin{tabular}{cccc}
\toprule
 & Aishell-1 & \multicolumn{2}{c}{LibriSpeech} \\
 \cmidrule(lr){3-4}
 & & Clean & Other \\
\midrule
Instruction 1 & 12.32& 47.57& 51.10\\
Instruction 2 & 34.61& 48.33& 65.06\\
Instruction 3 & 23.19& 37.05& 48.10\\
Instruction 4 & 12.13& 23.07& 17.18\\
without LLM & 4.73 & 3.35 & 8.77 \\
\bottomrule
\end{tabular}
\end{table}
\setlength{\tabcolsep}{13pt}

\subsection*{Results with Varying Shots}
\label{app_shot}
We evaluated the effect of varying the number of examples provided to the model, using 1-shot, 2-shot, and 3-shot configurations. We employed the GPT-3.5-turbo-16k-0613 model for this purpose. 
Tables \ref{tab:app_shots_1}, \ref{tab:app_shots_2}, and \ref{tab:app_shots_3} depict the experimental results using Instructions 1, 2, and 3, respectively.

Our findings suggest that providing more examples to the model leads to improved performance. This is consistent with results observed in numerous NLP tasks involving LLM. However, in the 1-shot, 2-shot, and 3-shot scenarios we tested, none yielded a satisfactory WER, indicating an increase in errors after correction. This aligns with our previous observation that additional efforts are required to effectively employ LLMs for ASR transcription error correction.

\begin{table}[H]
\centering
\caption{WER comparison for varying shots with Instruction 1 and the GPT-3.5-turbo-16k-0613 model.}
\label{tab:app_shots_1}
\begin{tabular}{cccc}
\toprule
 & Aishell-1 & \multicolumn{2}{c}{LibriSpeech} \\
 \cmidrule(lr){3-4}
 & & Clean & Other \\
\midrule
1-shot & 12.36& 47.93& 51.25\\
2-shot & 20.67& 70.93& 73.32\\
3-shot & 38.49& 81.43& 76.58\\
without LLM & 4.73 & 3.35 & 8.77 \\
\bottomrule
\end{tabular}
\end{table}

\begin{table}[H]
\centering
\caption{WER comparison for varying shots with Instruction 2 and the GPT-3.5-turbo-16k-0613 model.
}
\label{tab:app_shots_2}
\begin{tabular}{cccc}
\toprule
 & Aishell-1 & \multicolumn{2}{c}{LibriSpeech} \\
 \cmidrule(lr){3-4}
 & & Clean & Other \\
\midrule
1-shot & 34.08& 48.58& 64.60\\
2-shot & 45.70& 80.04& 94.20\\
3-shot & 76.48& 80.39& 90.79\\
without LLM & 4.73 & 3.35 & 8.77 \\
\bottomrule
\end{tabular}
\end{table}

\begin{table}[H]
\centering
\caption{WER comparison for varying shots with Instruction 3 and the GPT-3.5-turbo-16k-0613 model.}
\label{tab:app_shots_3}
\begin{tabular}{cccc}
\toprule
 & Aishell-1 & \multicolumn{2}{c}{LibriSpeech} \\
 \cmidrule(lr){3-4}
 & & Clean & Other \\
\midrule
1-shot & 22.32& 37.21& 48.14\\
2-shot & 52.52& 67.10& 72.88\\
3-shot & 86.49& 66.78& 69.46\\
without LLM & 4.73 & 3.35 & 8.77 \\
\bottomrule
\end{tabular}
\end{table}

\end{document}